\documentclass[11pt,letterpaper]{article}

\usepackage{setspace, enumitem, graphicx, amsmath, amssymb, sectsty, titlesec, booktabs, float, siunitx, subfigure, soul, xcolor, multirow, pbox}
\usepackage{float}
\usepackage{geometry}
\usepackage[numbers,sort&compress]{natbib}

\usepackage[activate={true,nocompatibility},final,tracking=true,kerning=true,spacing=true,factor=1100,stretch=10,shrink=10]{microtype}
\usepackage{booktabs}
\usepackage{array}
\usepackage{multirow}
\usepackage{xcolor}
\usepackage{colortbl}
\usepackage{tabularx}

\usepackage{hyperref}
\hypersetup{
	colorlinks = true, linkcolor = blue, anchorcolor = blue,
	citecolor = blue,  filecolor = blue, urlcolor = blue}

\usepackage[capitalise,nameinlink]{cleveref}
\usepackage{titlesec}

\usepackage[justification=centering]{caption}

\usepackage{booktabs}
\usepackage{array}
\usepackage{float}

\newcolumntype{L}[1]{>{\raggedright\arraybackslash}p{#1}}
\newcolumntype{C}[1]{>{\centering\arraybackslash}p{#1}}

\usepackage{fancyhdr}
\usepackage{cleveref}

\graphicspath{{files/}}

\sectionfont{\fontsize{12}{15}\selectfont}
\subsectionfont{\fontsize{12}{15}\selectfont}
\usepackage{fancyhdr}
\usepackage{fancyhdr}
\titlespacing{\section}{0pt}{4pt}{5pt}
\titleformat{\section}{\bfseries}{\thesection.}{0.5em}{}
\titleformat{\subsection}{\itshape}{\thesubsection.}{0.5em}{}
\titlespacing*{\section}{0pt}{1.0ex plus 1ex minus .8ex}{0.5ex}
\titlespacing*{\subsection}{0pt}{0.8ex plus .5ex minus .2ex}{0.3ex}

\makeatletter
\def\@maketitle{%
	\begin{center}%
		\textbf{\fontsize{13}{17}\selectfont \@title}
	\end{center}%
}
\makeatother

\usepackage{fancyhdr}
\begin{document}

\title{Data-Driven Modeling of Molten Chloride Corrosion Using Static Test Data}

\maketitle
\vspace{-18pt}
\begin{center}
    \text{Md Mahmudul Hasan Pathik$^{a,*}$}
\end{center}
\vspace{-20pt}
\begin{flushleft}
\begin{center}\small
\textit{$^{a}$Mechanics and Materials Lab, North Carolina State University, NC 27606 USA\\[0.10\baselineskip]}

\end{center}
\end{flushleft}

\setstretch{1.16}
\section*{Abstract}
Molten chloride corrosion has been studied for decades, yet reliable prediction across different alloys, salts, and test conditions remains difficult. In this work, 291 static molten chloride corrosion experiments reported in the literature were compiled into a structured dataset and used to develop an interpretable machine-learning framework for corrosion-rate predictions. To better represent the differences in salt condition across studies, a Chloride Salt Preparation Index (CSPI) was introduced to bring the reported salt-treatment method, atmosphere control, and pre-exposure impurity condition on a common numerical scale. A blended tree-based ensemble provided the most consistent predictive performance, achieving a cross-validated \(R^2\) of 0.85 and an MAE of 0.41 mg cm\(^{-2}\) day\(^{-1}\). The graded CSPI retained more useful information than a simple purified/unpurified classification and emerged as the dominant model descriptor. Global sensitivity and SHAP analyses further identified exposure time, temperature, and alloy chemistry as important contributors to the model predictions, with several observed trends consistent with the established molten-chloride corrosion behavior. The conformal prediction intervals were well calibrated, providing an estimate of prediction uncertainty in addition to the point prediction. Overall, the framework demonstrates that historical corrosion data can support quantitative, interpretable, and uncertainty-aware prediction when experimental context is represented explicitly, while also highlighting where more consistent reporting and targeted experiments are needed to improve future molten-salt corrosion models.

\section*{Keywords}
Molten salt corrosion, Static Corrosion Testing, Machine learning,
Salt purification, Uncertainty quantification.

\vspace{1cm}
\noindent\rule{9cm}{0.4pt} 

\vspace{0.2cm}
\noindent{\small$^{*}$ \textbf{Corresponding author}: }  

\noindent\textit{Email address}: \textit{mhpathik@ncsu.edu }.
\clearpage

\section{Introduction}
The effort for advanced energy systems has renewed interest in molten salt technologies, particularly for next-generation molten salt reactors (MSRs) and concentrated solar power (CSP) applications. These systems offer compelling advantages including high operating temperatures, excellent heat transfer characteristics, and inherent safety features due to some advantageous properties of molten salts, such as high boiling points, low vapor pressures, large
heat capacity, and thermal conductivity \cite{sohal2010engineering,janz1981physical}. The successful deployment and commercialization of MSR and CSP technologies largely depends on addressing the key challenge of structural component degradation in a high-temperature, corrosive molten salt environment. While high temperature alloys such as stainless steels (e.g., 316L, 316H), Ni-based and Fe-Ni-Cr high-temperature alloys (e.g., Alloy 617, Alloy 625, Alloy 800H) have been identified as primary candidates \cite{Dolan2024, Hartmann2022, SouthernCompany2022, SRNL2019, SRNL2020, Walter2022,Busby2019}, uncertainties remain regarding their long-term structural integrity and performance in molten salt.

Corrosion in molten salts involves simultaneous electrochemical and thermochemical processes such as dissolution, intergranular attack, and impurity-driven reactions, each influenced by alloy composition, salt chemistry, temperature gradients, and flow conditions \cite{sridharan2013corrosion}. Unlike high-temperature oxidation in gasses, molten-salt corrosion generally does not benefit from the formation of a stable, protective oxide scale. In chloride salts, corrosion products of many alloying elements can dissolve into the molten salt, allowing continued material loss without the development of a passive layer. Temperature and exposure duration further affect reaction kinetics, transport, and the evolution of the metal–salt interface.  Experimental choices such as salt preparation, atmosphere control, containment material, and the salt-volume-to-exposed-surface-area ratio ($V/SA$) can introduce additional variation. The measured corrosion response therefore reflects the combined influence of material composition, salt chemistry, operating conditions, and experimental practice. To understand all those, a
variety of test systems have been developed over the decades to evaluate material performance in
molten salt. Historically, the evaluation of material performance has relied on two primary methodologies: static crucible or capsule tests and flow circulation loops. From the 1950s through 1970s, ORNL investigated corrosion through numerous static crucible tests, thermal, and forced convection loop tests \cite{olson2009materials, ambrosek2011molten, ouyang2014long, gong2022molten, koger1972effect, koger1972evaluation, crowley1963design, silverman1976heat,kelleher2022thermal}. Static tests allow for primary evaluation of chemical interactions between the metal and salt at equilibrium or near-equilibrium conditions \cite{Raimanreport}, whereas loop tests additionally incorporate effects of flow, temperature gradients and mass transport. These experiments are expensive and time-intensive, often requiring long hours of continuous testing to generate statistically meaningful data. 

In parallel, physics-based models \cite{pillai2021first}, such as diffusion-controlled dissolution and thermodynamic activity-based approaches in corrosion modeling, have been developed to simulate observed degradation behavior and extend predictive capabilities beyond the constraints of laboratory testing. These approaches provide valuable mechanistic insight but typically require extensive parameterization and are constrained by incomplete knowledge of impurity effects, multi-component interactions, and long-term kinetic evolution. Rippy et al. \cite{rippy2023predicting}, for example, showed that the combined Ellingham and chloride–oxide stability diagrams can explain several experimentally observed trends in chloride corrosion. Such methods are particularly useful for determining whether a reaction is thermodynamically favorable and for interpreting the relative stability of alloy constituents. However, they are less suitable for directly predicting a quantitative corrosion rate across the broad range of conditions represented in the experimental literature. This study identified the absence of a comprehensive method for predicting corrosion in molten chloride salts, and subsequent reviews have continued to treat quantitative forecasting across different material–salt systems as an unresolved problem. As a result, quantitative rate predictions that generalize across alloy compositions, chloride salt chemistries, temperatures, and exposure times remain limited. 
So, there is a pressing need for advanced, data-driven modeling tools that can integrate experimental data with material and salt compositions and other driving factors to predict molten salt corrosion rates by learning complex relationships from experimental data to predict corrosion rates, identify critical operating parameters, and optimize alloy selection without large testing.

To better account for the coupled effects of corrosion variables, recent studies have increasingly turned to machine learning (ML) methods because of its ability to extract relationships from datasets containing multiple interacting variables and complex, nonlinear trends.  Corrosion prediction using neural networks dates back to the 1990s, when multilayer perceptrons and back-propagation networks were first applied to estimate localized corrosion and pitting behavior \cite{cai1999phenomenological,parthiban2005neural}. Since then, machine learning (ML) has evolved through more powerful algorithms, including support-vector regression (SVR), random-forest regression (RFR), gradient-boosting regression (GBR), and long-short-term-memory (LSTM) networks for modeling both static and time-series corrosion data \cite{muthukrishnan2024machine,coelho2022reviewing,diao2021improvement,wang2023machine}. Unlike purely physics-based models, which rely on predefined assumptions and simplified kinetics, ML approaches can capture highly nonlinear relationships among multiple variables such as temperature, alloy composition, salt chemistry, impurity levels, and flow conditions directly from experimental data \cite{yan2020corrosion}. ML therefore provides a useful framework for evaluating several material and environmental parameters simultaneously and determining which combinations of variables contain the greatest predictive information.

In this study, corrosion data reported in the literature for alloys exposed to static molten chloride salts are compiled into a structured database and analyzed using both statistical methods and machine-learning algorithms. A Chloride Salt Preparation Index (CSPI) is developed and introduced as a descriptor in the dataset to account for differences in salt purification process among literature corrosion studies. The effects of alloy composition, salt chemistry, temperature, exposure duration, purification conditions, and other experimental variables on corrosion behavior are systematically examined. The predictive capability of different ML models is then evaluated for corrosion-rate estimation, while feature-importance and sensitivity analysis are used to identify the factors that contribute most strongly to the predicted response. This study demonstrates the feasibility and advantages of applying machine learning for heterogeneous molten-salt corrosion data, identifies trends, quantifies prediction uncertainty, and guides future test selection to support more informed prediction.

\section{Methodology}
\subsection{Corrosion Data and Data Processing}
Reliable corrosion prediction using ML depends strongly on the availability of sufficiently large, high-quality datasets with consistent and well-defined features. A substantial amount of experimental data has been generated over decades; much of it originates from laboratory-scale studies rather than operational experience. The interpretation of these results is often complicated by variations in test methods, alloy compositions, salt chemistry, and environmental conditions, compounded by the absence of standardized corrosion testing methods for molten salt systems \cite{epri3002030754}. Salt preparation introduces another important source of variability. Some studies rely primarily on thermal purification, whereas others employ additional chemical purification steps, which can alter impurity concentrations and consequently affect the measured corrosion response \cite{park2024static}. All of these differences contribute additional uncertainty and complicate cross-study comparison. 

In this study, a structured dataset was developed from the literature on molten chloride salt corrosion experiments \cite{osti_4023508, ambrosek2010molten, abramov2014corrosion, polovov2017corrosion, ding2018hot, sun2018corrosion, sun2018effects, ai2023study, kurley2019enabling,
phillips2019corrosion, cho2019hot, raiman2019compatibility, guo2020excellent, knosalla2020investigation, Dsouza2021MaterialDegradation, dsouza2021corrosion,
hamdy2021perspectives, li2021effect, ren2021corrosion, dingreville2021multi, gong2022molten, mortazavi2022high, xu2022high, keiser2023material, Pint2023StaticSaltCompatibility, Lee2023metalassisted, lu2023excellent, li2023effect,gong2023molten, wang2024high, kondaiah2024electrodeposited, okonkwo2024corrosion,
chi2024effects, choi2024corrosion, duan2024comparative, yan2024corrosion, yang2024hot, lv2025corrosion, li2025investigation, pillai2025alloy244,
xue2026enhancing}. The present dataset was compiled to include parameters that describe both material and salt characteristics relevant to corrosion under static conditions. Only chloride-based salts were considered in the present dataset to maintain consistency in the underlying corrosion chemistry, as fluoride and nitrate salts can exhibit substantially different corrosion mechanisms and interfacial reactions \cite{ma2022comparative}. The final dataset contains 291 individual observations, each representing a unique experimental condition. It includes alloying elements of corrosion coupons, processing parameters (e.g. temperature, exposure duration, and CSPI) and compounds present or added in the salts. In static testing, container (capsule/crucible) material can significantly affect the measured corrosion rate of the coupon through cross-contamination, galvanic coupling, or element exchange \cite{park2024static,zhengetal2015corrosion,massimi2024assessment}. Differences in chemical potential between coupon and container drive mass transfer of reactive species (especially Cr), altering both local redox potential and alloy composition \cite{falconer2022activity}. Hence, container elements are also included alongside coupon elements to capture such interfacial coupling effects between the specimen and its environment.

\begin{figure}[H]
    \centering
{\includegraphics[width=0.99\linewidth]{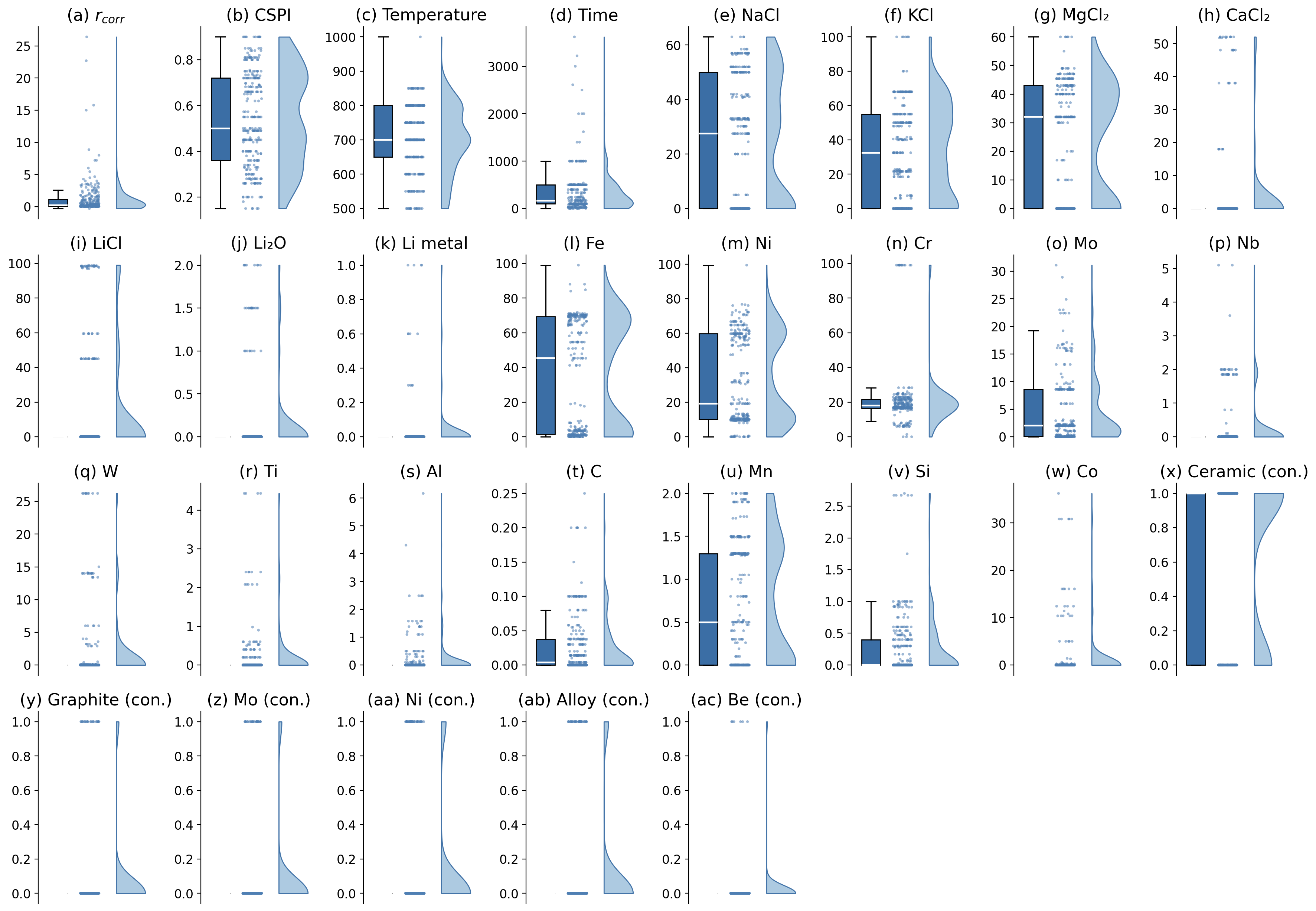}}

    \caption{The raw data distribution for each variable in chloride salt is shown using a box plot (left), a jittered scatter plot (center), and a kernel density curve (right) within each subplot.}
    \label{fig:fg1}
   \end{figure}

The statistical distribution of each active feature for chloride salt system is shown in \cref{fig:fg1}  using a combined box-strip-density representation, where the box plot captures the interquartile range and median, the strip plot shows individual data points with horizontal jitter to reveal clustering patterns, and the kernel density estimate provides a smoothed approximation of the underlying probability distribution. The dataset contains 29 numerical variables, of which 28 are used as model inputs and the corrosion-rate ($r_{corr}$) as the target. The input features include 12 alloying-element descriptors, six container-material (cont.) descriptors, seven salt-phase constituents and impurities, and three processing parameters: CSPI, temperature, and exposure time. These variables were retained during feature selection to evaluate their individual and combined contributions to corrosion behavior.
The salt composition features reflect the chloride salt systems investigated in the literature, with KCl, MgCl$_2$, NaCl, LiCl and CaCl$_{2}$  representing the primary constituents. Notably, two salt-phase additives Li and Li$_{2}$O were present in fewer than 20\% of the observations. All descriptors are listed with their ranges, mean, and standard deviation (S.D) in \cref{tab:chloride_inputs}. The majority of coupon materials are mostly Fe-based ( e.g. SS 304, 316H, 321, etc.) and Ni-based alloys ( Alloy 600, 617, 625, 800H, 718 etc.) . In contrast to the coupon, the container is encoded by material identity rather than by elemental composition, using six binary indicators such as ceramic (alumina or quartz), graphite, molybdenum, nickel, structural alloy, and beryllium. Ceramic containment dominates the compilation at samples (62.5\%), followed by nickel (13.4\%), structural alloy (13.1\%), molybdenum (8.9\%), graphite (7.9\%), and beryllium (2.1\%). Elemental container compositions were not compiled, since approximately 70\% of the experiments used ceramic or graphite containment, for which metallic compositions are inapplicable.
\begin{table}[H]
\centering
\caption{Input features and ranges for the chloride-salt dataset ($N=291$).}
\label{tab:chloride_inputs}
\small
\setlength{\tabcolsep}{6pt}
\begin{tabular}{l l c c c c}
\toprule
\textbf{Category} & \textbf{Element or detail} & \textbf{Unit} & \textbf{Range} & \textbf{Mean} & \textbf{S.D} \\
\midrule
\multirow{12}{*}{\shortstack[l]{Coupon\\ elements}}
 & Fe & wt.\% & 0 -- 99.0   & 35.82 & 32.04 \\
 & Ni & wt.\% & 0 -- 99.4   & 32.36 & 25.31 \\
 & Cr & wt.\% & 0 -- 99.0   & 21.27 & 17.60 \\
 & Mo & wt.\% & 0 -- 31.1   & 4.32  & 5.96 \\
 & Nb & wt.\% & 0 -- 5.10   & 0.25  & 0.74 \\
 & W  & wt.\% & 0 -- 26.2   & 1.86  & 5.50 \\
 & Ti & wt.\% & 0 -- 4.43   & 0.20  & 0.66 \\
 & Al & wt.\% & 0 -- 6.17   & 0.18  & 0.61 \\
 & C  & wt.\% & 0 -- 0.25   & 0.025 & 0.042 \\
 & Mn & wt.\% & 0 -- 2.00   & 0.70  & 0.72 \\
 & Si & wt.\% & 0 -- 2.70   & 0.26  & 0.47 \\
 & Co & wt.\% & 0 -- 36.2  & 1.47  & 5.33 \\
\midrule
\multirow{7}{*}{\shortstack[l]{Salt\\ compounds}}
 & NaCl      & mol.\% & 0 -- 63.0  & 23.27 & 22.95 \\
 & KCl       & mol.\% & 0 -- 100.0 & 33.26 & 27.60 \\
 & MgCl$_2$  & mol.\% & 0 -- 60.0  & 23.60 & 20.21 \\
 & CaCl$_2$  & mol.\% & 0 -- 52.0  & 4.29  & 13.32 \\
 & LiCl      & mol.\% & 0 -- 99.0  & 15.40 & 32.30 \\
 & Li$_2$O   & mol.\% & 0 -- 2.00  & 0.16  & 0.48 \\
 & Li (metal)& mol.\% & 0 -- 1.00  & 0.03  & 0.14 \\
\midrule
\multirow{3}{*}{\shortstack[l]{Processing \\ conditions}}
 & CSPI score  & --   & 0.15 -- 0.90    & 0.54   & 0.21 \\
 & Temperature & \si{\celsius} & 500 -- 1000 & 698.97 & 96.13 \\
 & Time        & h    & 1 -- 3620       & 343.96 & 489.76 \\
\midrule
\multirow{6}{*}{\shortstack[l]{Container\\ type}}
 & Ceramic  & - & 0 -- 1 & 0.625 & 0.485 \\
 & Graphite & - & 0 -- 1 & 0.079 & 0.270 \\
 & Mo       & - & 0 -- 1 & 0.089 & 0.286 \\
 & Ni       & - & 0 -- 1 & 0.134 & 0.341 \\
 & Alloy    & - & 0 -- 1 & 0.131 & 0.338 \\
 & Be       & - & 0 -- 1 & 0.021 & 0.142 \\
\midrule
Target property & Corrosion rate & mg\,cm$^{-2}$\,day$^{-1}$ & $-0.29$ -- 26.40 & 1.124 & 2.679 \\
\bottomrule
\end{tabular}
\end{table}

The test conditions are described by test temperature, exposure duration, and a graded salt-preparation score that was constructed for the present work by rating the purification procedure described in each source publication to capture how salt purification influences corrosion severity in molten‑salt systems. Salt preparation can substantially alter impurity concentrations and therefore the measured corrosion response. Thus, encoding salt preparation and purification on a continuous scale (0–1) allows the model to learn its correlation with corrosion rate across diverse molten‑salt experiments.

The target feature considered in this study is corrosion rate, $r_{corr}$ (mg\,cm$^{-2}$\,day$^{-1}$) during exposure to molten salt. Other potential corrosion metrics, such as penetration depth and chromium depletion, were not included because consistent and sufficient data were not available in the published literature for chloride salt systems. The reported corrosion rate data are highly skewed, with most values concentrated near zero and higher values reaching up to 26.4 mg\,cm$^{-2}$\,day$^{-1}$ as mentioned in \cref{tab:chloride_inputs}. In addition, negative corrosion values (corresponding to mass gain) are present in some datasets, reflecting oxidation or deposition phenomena rather than material loss.

\begin{figure}[H]
    \centering
    \subfigure[]{\includegraphics[width=0.46\linewidth]{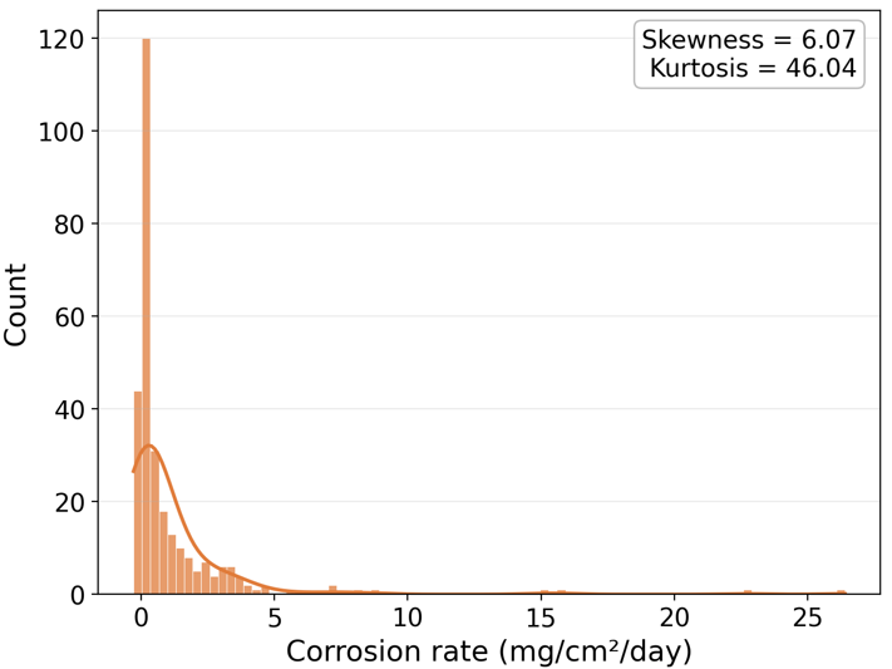}}
	\subfigure[]{\includegraphics[width=0.452\linewidth, height=5.8cm]{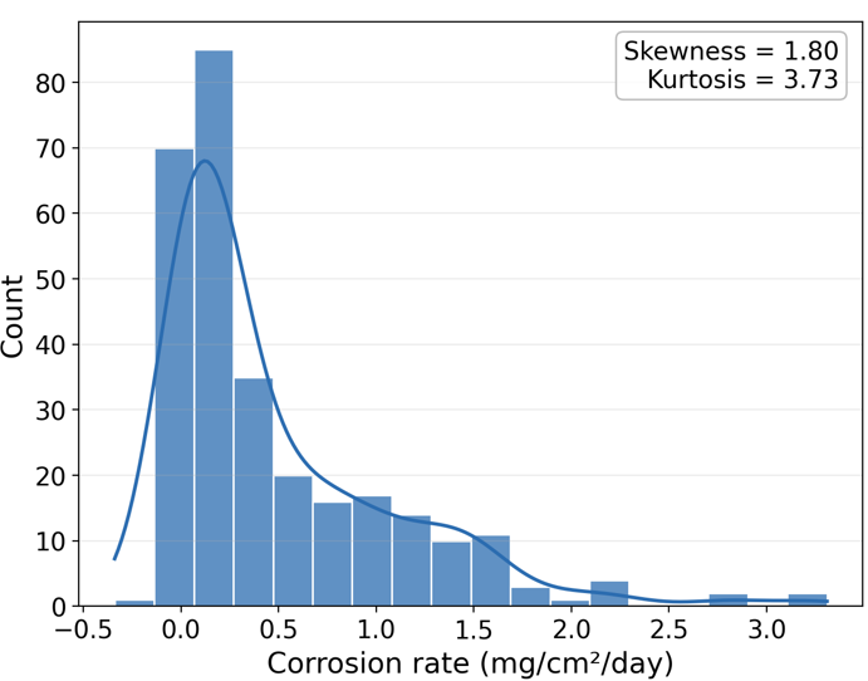}} 
    \caption{Distributions of corrosion rates (a) before and (b) after log transformation.}
    \label{fig:fg2}
   \end{figure}

To address this heteroskedasticity, a logarithmic transformation was applied to the
target variable according to \cref{eq:log_transform}:
\begin{equation}
    y' = \ln\left(1 + y\right)
    \label{eq:log_transform}
\end{equation}
where $y$ is the raw corrosion rate in mg\,cm$^{-2}$\,day$^{-1}$. The transformation reduces the skewness of the target from 6.07 to 1.8
(\cref{fig:fg2}) and stabilizes the variance \cite{wand1991transformations}, so that
the regression algorithms operate in a logarithmic space and penalize relative
rather than absolute errors. Because the transformation contains no fitted
parameter, it is identical in every training and test partition and does not
transfer information from held-out records. Predictions were returned to the
original scale as $\hat{y} = \exp(\hat{y}') - 1$.

\begin{figure}[H]
    \centering
    \includegraphics[width=.95\linewidth, height=18.0cm, keepaspectratio]{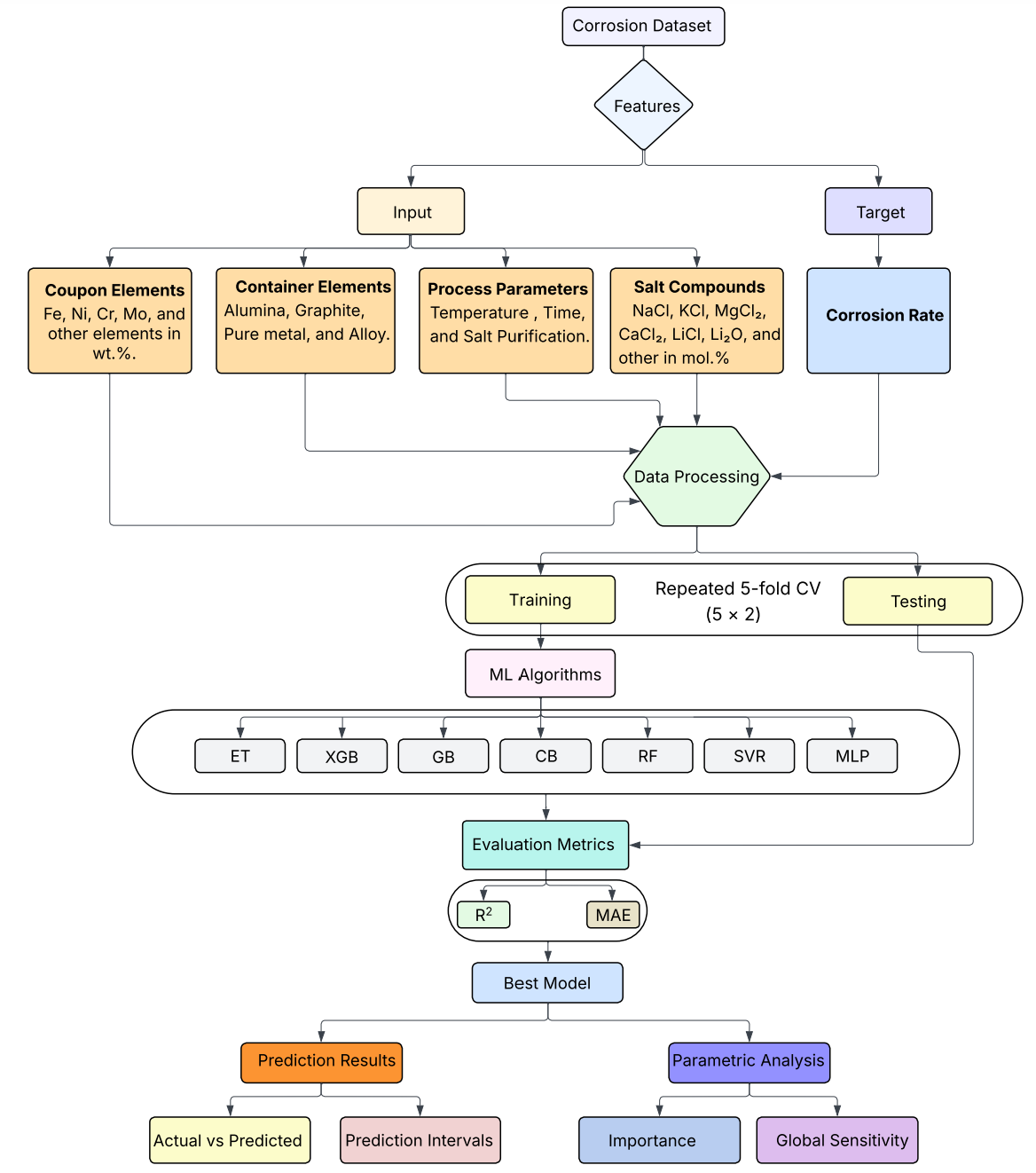}
    \caption{Overview of the machine learning workflow used to predict corrosion rates in molten salt systems.}
    \label{fig:fg3}
\end{figure}

Following data processing, all models were evaluated, multiple machine-learning algorithms are then trained and evaluated using the coefficient of determination ($R^2$) and mean absolute error (MAE). The selected model is subsequently used to compare predicted and measured corrosion rates, examine prediction uncertainty, and global feature sensitivity. The overall modeling workflow is summarized in \cref{fig:fg3}.

\subsection{Machine Learning Model Construction and Evaluation}
To capture a range of nonlinear relationships within the data, we evaluated seven regression algorithms spanning kernel, bagging, boosting, and neural-network families. These included Support Vector Regression (SVR) with a Radial Basis Function (RBF) kernel, Random Forest (RF), Extra Trees (ET), Gradient Boosting Regressor (GBR), Extreme Gradient Boosting (XGB), CatBoost and a Multilayer Perceptron (MLP).

\noindent For each algorithm, hyperparameter tuning was conducted as shown in \cref{tab:hyperparams}, with configurations selected based on cross-validated training performance. We also evaluated a blended ensemble consisting of four learners: XGB, CatBoost, ET, and GBR trained on the logarithmically transformed target. The four predictions were then averaged, and the resulting ensemble prediction was transformed back to the original corrosion-rate scale. Equal weights were assigned to the four models because estimating separate model weights from only ten cross-validation evaluations could introduce additional overfitting and model-selection uncertainty.

Model performance was evaluated using five-fold cross-validation with two repetitions, resulting in ten held-out evaluations for each model. The same fold assignments were used for every model so that their performances could be compared on identical training and testing observations. Because identical cross-validation folds were used for all configurations, model comparisons were performed on paired fold-level results. Feature scaling was fitted on each training fold only and then applied to the corresponding held-out fold. Because the hyper-parameters in \cref{tab:hyperparams} were selected on the same compilation, their effect on the reported performance was estimated by nested cross-validation, in which each member's configuration was selected by an inner three-fold cross-validation on the training portion of each outer fold. Model performance was evaluated using the evaluation metrics in \cref{eq:r2,eq:mae}, and the values for each model are summarized in \cref{tab:train_test}.

\begin{equation}
R^2=1-\frac{\sum_{}(y_i-f_i)^2}{\sum(y_i-\bar{y})^2}
\label{eq:r2}
\end{equation}

\begin{equation}
\mathrm{MAE}=\frac{1}{n}\sum_{i=1}^{n}\lvert y_i-f_i\rvert
\label{eq:mae}
\end{equation}
where $f_i$ is the predicted corrosion rate, $y_i$ is the corresponding measured value, $\bar{y}$ is the mean measured corrosion rate, and $n$ is the number of observations in the evaluated fold. Because predictions were back-transformed before evaluation, MAE is reported in mg\,cm$^{-2}$\,day$^{-1}$.

The cross-validated results in \cref{tab:train_test} show that the tree-based boosting models performed better than the kernel and neural-network models for the present dataset. XGB achieved the strongest performance among individual learners, with a validation $R^2$ of $0.81 \pm 0.04$ and an
MAE of $0.44 \pm 0.08$ mg\,cm$^{-2}$\,day$^{-1}$. The blended ensemble retained the same mean $R^2$ while slightly reducing the MAE to $0.43 \pm 0.08$ mg\,cm$^{-2}$\,day$^{-1}$. The ensemble was therefore selected as the reference model for subsequent feature engineering, sensitivity, and interpretation analyses.
\begin{table}[H]
\centering
\caption{Hyperparameter configurations for machine-learning models.}
\label{tab:hyperparams}
\small
\begin{tabular}{l l c c}
\toprule
\textbf{Model} & \textbf{Hyperparameter} & \textbf{Search range} & \textbf{Optimized value} \\
\midrule
\multirow{9}{*}{\shortstack[l]{XGBoost\\ \scriptsize}}
 & n\_estimators      & 200--1600      & 1582 \\
 & max\_depth         & 2--8           & 6 \\
 & learning\_rate     & 0.005--0.25    & 0.024 \\
 & subsample          & 0.5--1.0       & 0.502 \\
 & colsample\_bytree  & 0.4--1.0       & 0.748 \\
 & min\_child\_weight & 1--7           & 1 \\
 & reg\_alpha         & 0.001--5       & 0.086 \\
 & reg\_lambda        & 0.01--10       & 1.573 \\
 & gamma              & 0.0001--2      & 0.0015 \\
\midrule
\multirow{7}{*}{\shortstack[l]{CatBoost\\ \scriptsize}}
 & iterations         & 300--1200      & 1079 \\
 & depth              & 3--8           & 6 \\
 & learning\_rate     & 0.01--0.2      & 0.095 \\
 & l2\_leaf\_reg      & 0.5--15        & 2.151 \\
 & subsample          & 0.6--1.0       & 0.641 \\
 & random\_strength   & 0.001--5       & 0.618 \\
 & bagging\_temperature & 0--1         & 0.867 \\
\midrule
\multirow{5}{*}{\shortstack[l]{Extra Trees\\ \scriptsize}}
 & n\_estimators      & 300--1200      & 404 \\
 & max\_depth         & \{None,10,20,30,40\} & 20 \\
 & min\_samples\_split& 2--11          & 3 \\
 & min\_samples\_leaf & 1--7           & 1 \\
 & max\_features      & 0.3--1.0       & 0.698 \\
\midrule
\multirow{7}{*}{\shortstack[l]{Gradient Boosting\\ \scriptsize}}
 & n\_estimators      & 200--1000      & 999 \\
 & max\_depth         & 2--6           & 5 \\
 & learning\_rate     & 0.005--0.25    & 0.112 \\
 & subsample          & 0.6--1.0       & 0.831 \\
 & min\_samples\_split& 2--11          & 3 \\
 & min\_samples\_leaf & 1--7           & 7 \\
 & max\_features      & 0.4--1.0       & 0.932 \\
\midrule
\multirow{5}{*}{\shortstack[l]{Random Forest\\ \scriptsize}}
 & n\_estimators      & 200--1000      & 371 \\
 & max\_depth         & \{None,10,20,30,40\} & 30 \\
 & min\_samples\_split& 2--11          & 2 \\
 & min\_samples\_leaf & 1--7           & 1 \\
 & max\_features      & 0.3--1.0       & 0.317 \\
\midrule
\multirow{3}{*}{\shortstack[l]{SVR (RBF)\\ \scriptsize}}
 & C                  & 0.1--100       & 12.31 \\
 & gamma              & 0.001--1       & 0.429 \\
 & epsilon            & 0.001--0.3     & 0.036 \\
\midrule
\multirow{4}{*}{\shortstack[l]{MLP\\ \scriptsize}}
 & hidden\_layer\_sizes & \{(32),(64),(64,32),(128,64),\dots\} & (64, 32) \\
 & activation         & \{relu, tanh\} & relu \\
 & alpha              & 0.0001--1      & 0.154 \\
 & learning\_rate\_init & 0.0001--0.01 & 0.0036 \\
\bottomrule
\end{tabular}
\end{table}

\begin{table}[H]
\centering
\caption{Cross-validated performance of the evaluated models using the
28 baseline descriptors. Values are reported as mean $\pm$ standard deviation
across ten held-out folds.}
\label{tab:train_test}
\begin{tabular}{l c c c}
\toprule
\textbf{Model} & \textbf{Train R$^2$} & \textbf{Test R$^2$} & \textbf{Test MAE} \\
\midrule
\textbf{Ensemble} & \textbf{0.98 } & \textbf{0.81 $\pm$ 0.04} & \textbf{0.43 $\pm$0.08} \\
XGB          & 0.98  & 0.81 $\pm$ 0.04 & 0.44 $\pm$ 0.08 \\
CB     & 0.98  & 0.76 $\pm$ 0.07 & 0.46 $\pm$ 0.11 \\
GBR          & 0.97  & 0.75 $\pm$ 0.11 & 0.47 $\pm$ 0.15 \\
ET           & 0.94  & 0.64 $\pm$ 0.29 & 0.50 $\pm$ 0.17 \\
RF           & 0.90  & 0.71 $\pm$ 0.11 & 0.56 $\pm$ 0.21 \\
SVR (RBF)    & 0.87  & 0.62 $\pm$ 0.16 & 0.58 $\pm$ 0.11 \\
MLP          & 0.57  & 0.54 $\pm$ 0.13 & 0.73 $\pm$ 0.25 \\
\bottomrule
\end{tabular}
\end{table}

\subsection{Feature Engineering}
The performance of the baseline model, which uses all 28 descriptors, is shown in \cref{tab:train_test}. This section evaluates whether comparable predictive performance can be achieved with fewer descriptors and whether corrosion-informed descriptors provide additional information beyond the original variables.

The blended ensemble was refitted for every candidate set, and every set was evaluated on the same ten folds, generated once by five-fold cross-validation repeated twice and then reused. Significance was assessed with the Wilcoxon signed-rank test and a t-test corrected for the overlap between training sets in repeated cross-validation. Because a ranking computed using all records could favor descriptors that happen to perform well on the held-out data, feature ranking and selection were repeated independently within each training fold. The ranking was based only on the training data, and the held-out fold was used solely to evaluate the ensemble refitted on the descriptors selected within that fold.

\begin{figure}[H]
    \centering
 \subfigure[] {\includegraphics[width=0.5\linewidth]{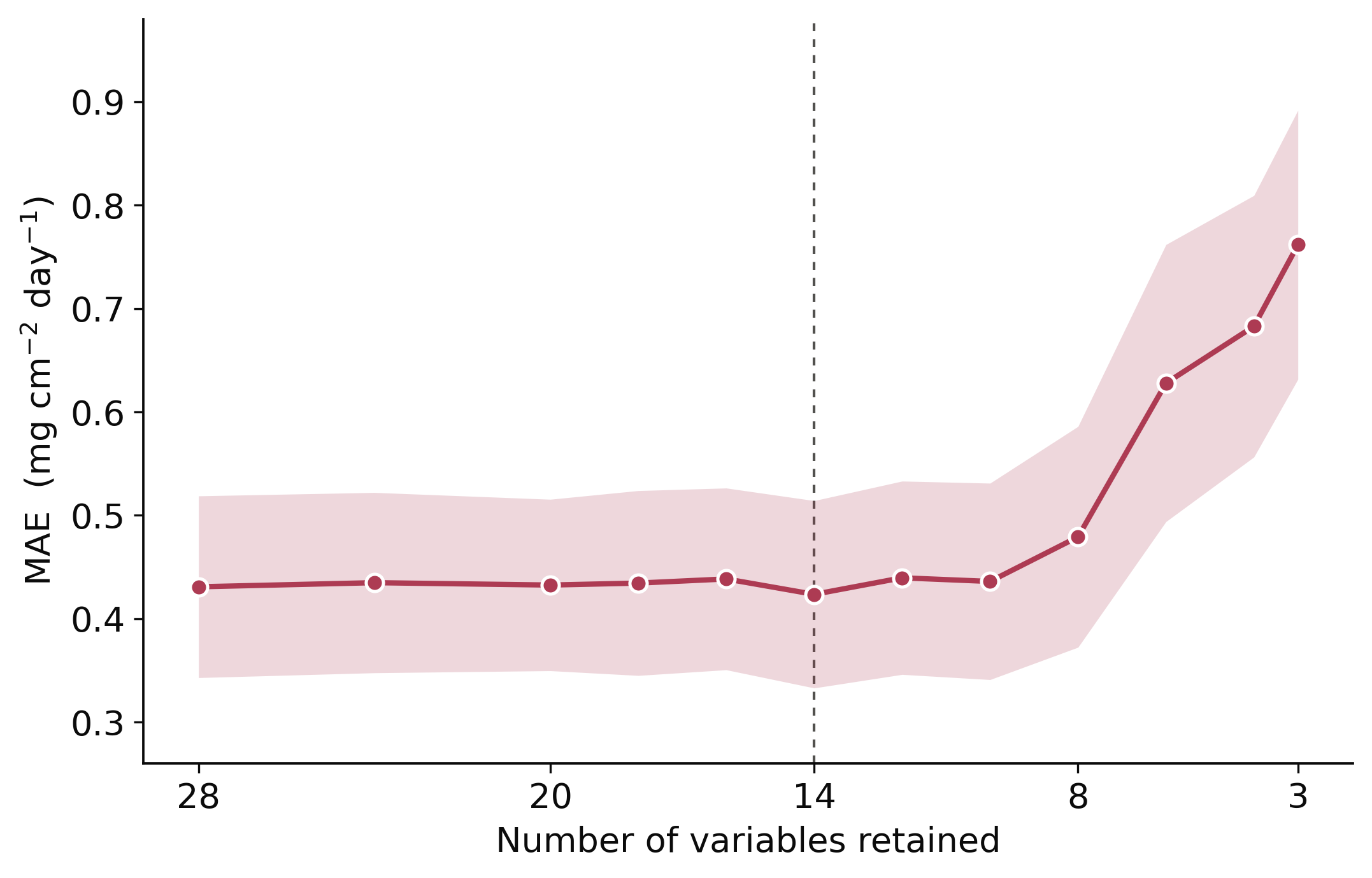}}
 \subfigure[] {\includegraphics[width=0.45\linewidth]{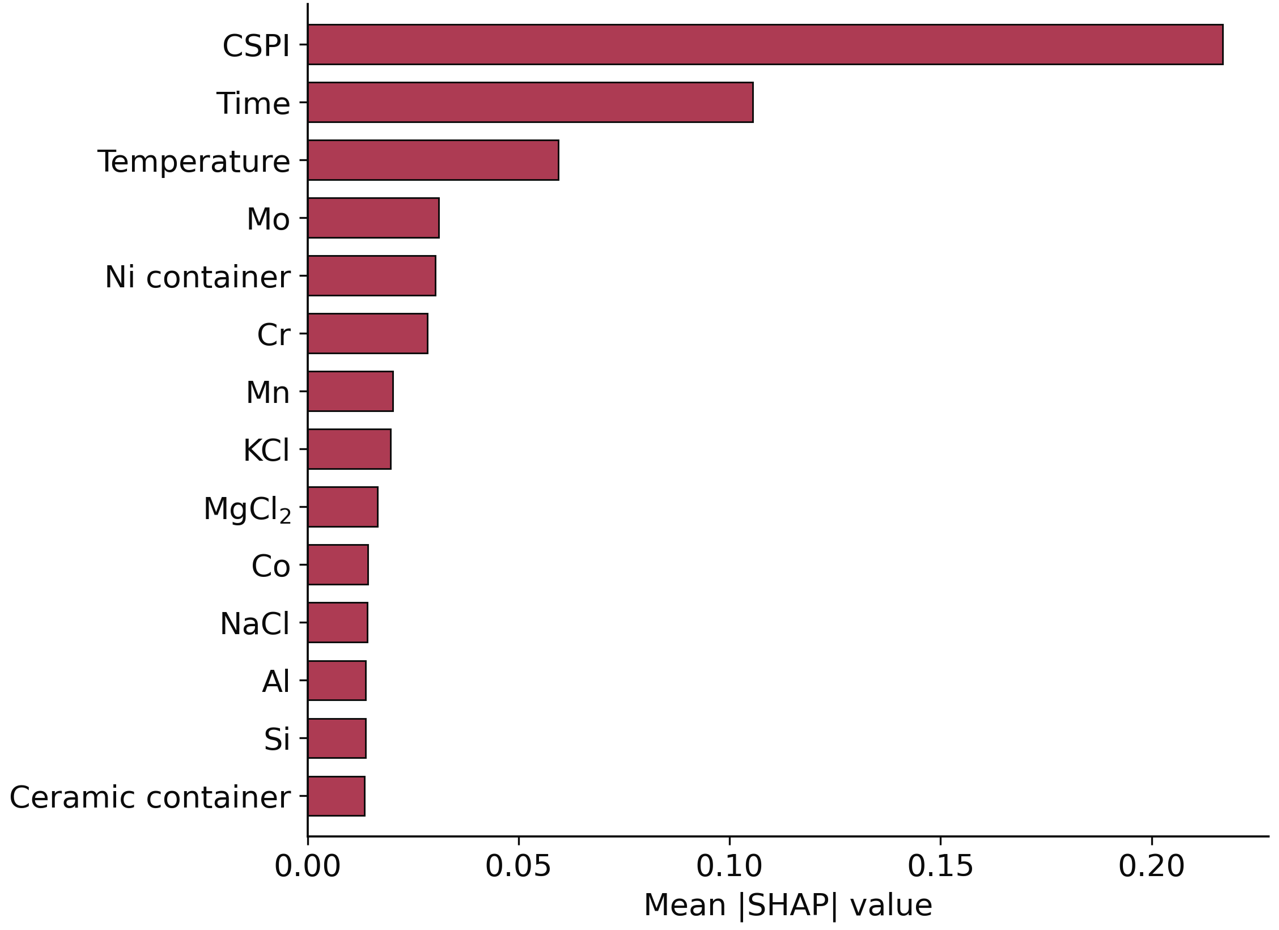}}
    \caption{ (a) Model performance during feature reduction and (b) importance of the final retained features.}
    \label{fig:fg4}
   \end{figure}

The variables were ordered by importance before any were removed and normalized to the highest value. The variables were progressively removed from the lower end of the importance ranking, and the ensemble was refitted after each removal (\cref{fig:fg4}). Model performance remained essentially unchanged when the number of variables was reduced from 28 to approximately 14. Further reduction caused a gradual increase in MAE, followed by a sharp rise when fewer than eight variables were retained as seen in \cref{fig:fg4} (a).

The 14 variables were ordered by importance as shown in \cref{fig:fg4} (b). Importance was taken as the mean absolute SHAP contribution of each variable, averaged over the ensemble members and computed out of fold. CSPI was by far the most influential variable, followed by exposure time  and temperature. All other variables contributed less than 0.035. The three test-condition variables dominate the SHAP-based feature ranking. Among the composition variables, Mo and Cr rank highest, consistent with the roles of the two elements in chloride melts.

\subsection{Effect of salt volume-to-surface-area ratio (V/SA) as an additional model input}

Salt \(V/SA\) can influence corrosion rate in static tests. At low \(V/SA\), the limited salt volume may become saturated with dissolved corrosion products, reducing the driving force for further dissolution. Higher \(V/SA\) generally maintains this driving force for longer and can produce greater mass loss and attack depth \cite{Pint2023StaticSaltCompatibility}. Therefore, differences in \(V/SA\) may partly explain variations among corrosion rates reported under otherwise similar conditions.

\begin{figure}[H]
    \centering
   {\includegraphics[width=0.78\linewidth]{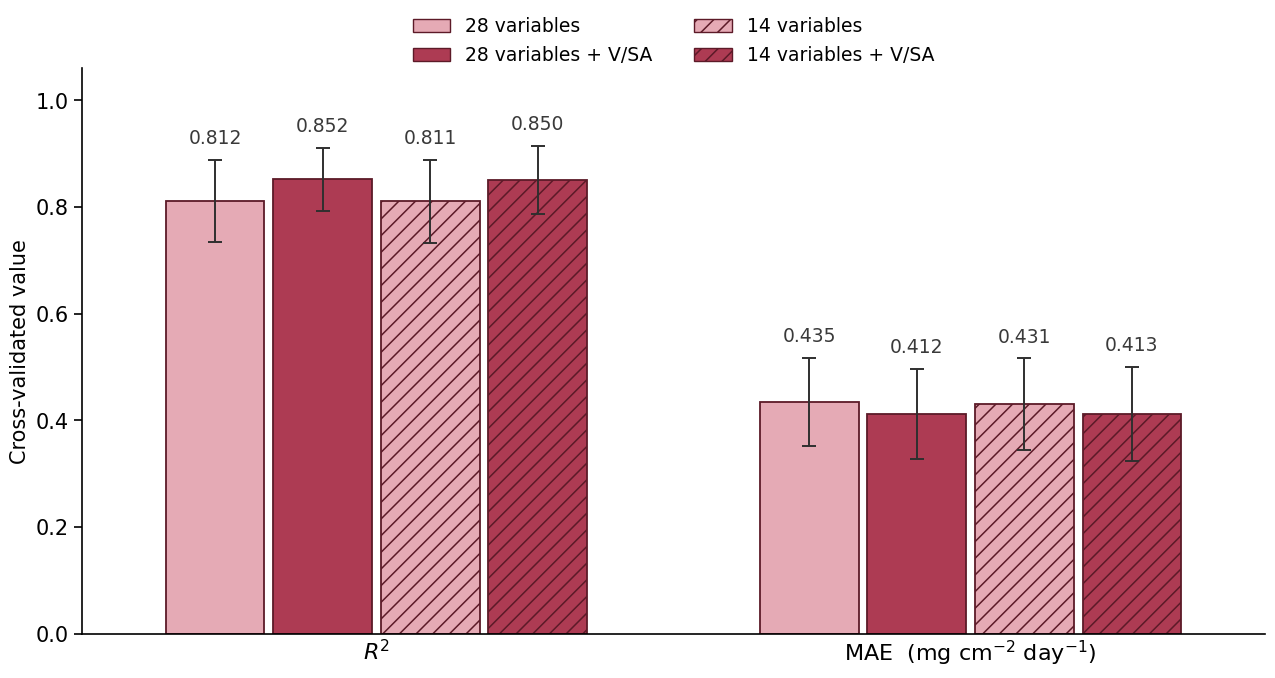}} 
    \caption{Influence of the V/SA on model performance, showing the cross-validated \(R^2\) and MAE for models with and without \(V/SA\).}
    \label{fig:fg5}
   \end{figure}

The initial baseline model used 28 input variables and did not include \(V/SA\). This parameter was available for only 160 of the 291 records. These reported values were added as a twenty-ninth input. The effect of adding  V/SA on model performance is shown in \cref{fig:fg5}.
 Including \(V/SA\) increased \(R^2\) from 0.81 to 0.85 and reduced MAE from 0.435 to 0.412 mg cm\(^{-2}\) day\(^{-1}\). The improvement in \(R^2\) occurred nine out of ten folds. Shuffling the reported values did not show measurable improvement, indicating that the numerical magnitude of \(V/SA\), rather than its reporting pattern alone, contributed to the prediction. Model performance remains unchanged as the set was reduced from 28 to 14 variables as seen in \cref{fig:fg5}.

\subsection{Conformal Prediction and Uncertainty Quantification}
\label{subsection:2.5}

The prediction intervals were constructed using conformal prediction to provide a measure of uncertainty in addition to the point prediction. The procedure was applied to the out-of-fold predictions generated during the repeated cross-validation, so that the residual associated with each observation was obtained from a model that had not been trained on that observation.

Let $z_i$ denote the transformed corrosion-rate target and $\hat{z}_i$ the corresponding out-of-fold ensemble prediction. For the standard conformal procedure, the nonconformity score was defined as the absolute prediction residual, as given in \cref{eq:eq4}.

\begin{equation}
\label{eq:eq4}
    s_i = |z_i-\hat{z}_i|.
\end{equation}

For a nominal coverage level of $1-\alpha$,  where $\alpha$ is the miscoverage level, the conformal half-width was obtained from the $(1-\alpha)$ quantile of the calibration nonconformity scores. The resulting prediction interval for a new observation with predicted value $\hat{z}_{\mathrm{new}}$ is given in \cref{eq:eq5}.

\begin{equation}
\label{eq:eq5}
    \left[
    \hat{z}_{\mathrm{new}}-q_{1-\alpha},
    \hat{z}_{\mathrm{new}}+q_{1-\alpha}
    \right],
\end{equation}

\noindent where $q_{1-\alpha}$ is the corresponding quantile of the calibration nonconformity scores.

The prediction error may vary across the input space; a normalized nonconformity score was also evaluated using disagreement among the four ensemble members. The ensemble disagreement for observation $i$, defined as the standard deviation of individual XGB, CatBoost, ET, and GBR predictions, is given in \cref{eq:eq6}.

\begin{equation}
\label{eq:eq6}
    \sigma_i =
    \operatorname{SD}
    \left(
    \hat{z}_{i,\mathrm{XGB}},
    \hat{z}_{i,\mathrm{CB}},
    \hat{z}_{i,\mathrm{ET}},
    \hat{z}_{i,\mathrm{GBR}}
    \right).
\end{equation}

Using this ensemble disagreement, the normalized nonconformity score is defined in \cref{eq:eq7}.

\begin{equation}
\label{eq:eq7}
    s_i^{*} =
    \frac{|z_i-\hat{z}_i|}
    {\sigma_i+\beta},
\end{equation}

where $\beta$, set to the median of $\sigma_i$, stabilizes the score for observations on which the ensemble members agree closely. With $\hat{q}^{*}_{1-\alpha}$ obtained from the normalized nonconformity scores, the corresponding normalized prediction interval is given in \cref{eq:eq8} which widens where the members disagree and narrows where they agree.

\begin{equation}
\label{eq:eq8}
    \left[
    \hat{z}_{\mathrm{new}}
    -\hat{q}^{*}_{1-\alpha}(\sigma_{\mathrm{new}}+\beta),
    \;
    \hat{z}_{\mathrm{new}}
    +\hat{q}^{*}_{1-\alpha}(\sigma_{\mathrm{new}}+\beta)
    \right].
\end{equation}

Calibration was evaluated on all 291 out-of-fold residuals. The scores were divided
into two halves, one used to set $\hat{q}$ and the other to measure empirical coverage, and the split was repeated 600 times; coverage and mean interval width were recorded over nominal levels from 0.50 to 0.99. These calibration splits were used to assess whether the conformal intervals achieved the intended coverage while maintaining practical interval widths across the evaluated confidence levels in the following section.

\section{Results and Discussion}
\subsection{Prediction Performance and Uncertainty Quantification} 
\begin{figure}[H]
    \centering
    \includegraphics[width=0.99\linewidth, keepaspectratio]{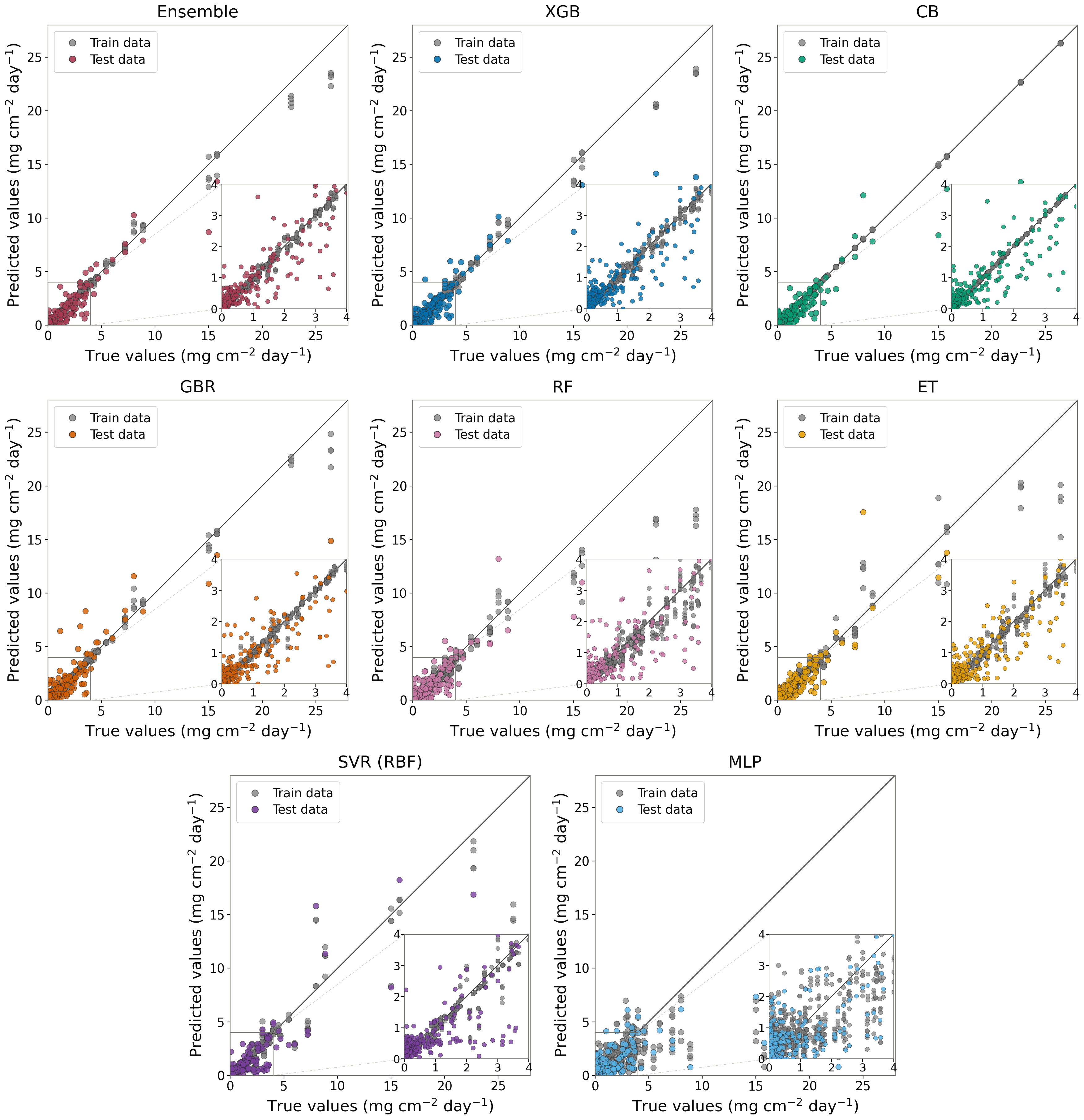} 
    \caption{Measured versus predicted corrosion rates for the evaluated models, with the insets highlighting the low-corrosion-rate region.}
    \label{fig:fg6}
   \end{figure}

The predictive behavior of the evaluated models is compared using the parity plots in \cref{fig:fg6}. The tree-based
models generally reproduce the measured response more closely than SVR and MLP, consistent with their stronger cross-validated performance. In particular, the ensemble and XGB predictions remain concentrated around the parity line
over most of the corrosion-rate range, while CB and GBR also capture the overall trend but show greater scatter among the held-out observations. RF and ET exhibit a wider spread from the parity line, whereas SVR and especially MLP show noticeably larger prediction errors across the test data.

The insets in \cref{fig:fg6} highlight the densely populated region below approximately 4~mg\,cm$^{-2}$\,day$^{-1}$. Although most observations occur
within this relatively low corrosion-rate range, considerable variability is still present among nominally similar rates. The ensemble and XGB maintain the closest clustering around the parity line in this region, whereas the remaining
models show progressively greater dispersion. 

At higher measured corrosion rates, the predictions become less uniform, and several models tend to underestimate the largest responses. This behavior is particularly evident for RF, ET, SVR, and MLP, while the ensemble remains relatively stable. The increased deviation at the upper end is consistent with the relatively sparse representation of severe-corrosion conditions in
the compiled literature dataset. 

\begin{figure}[H]
    \centering
    {\includegraphics[width=0.85\linewidth]{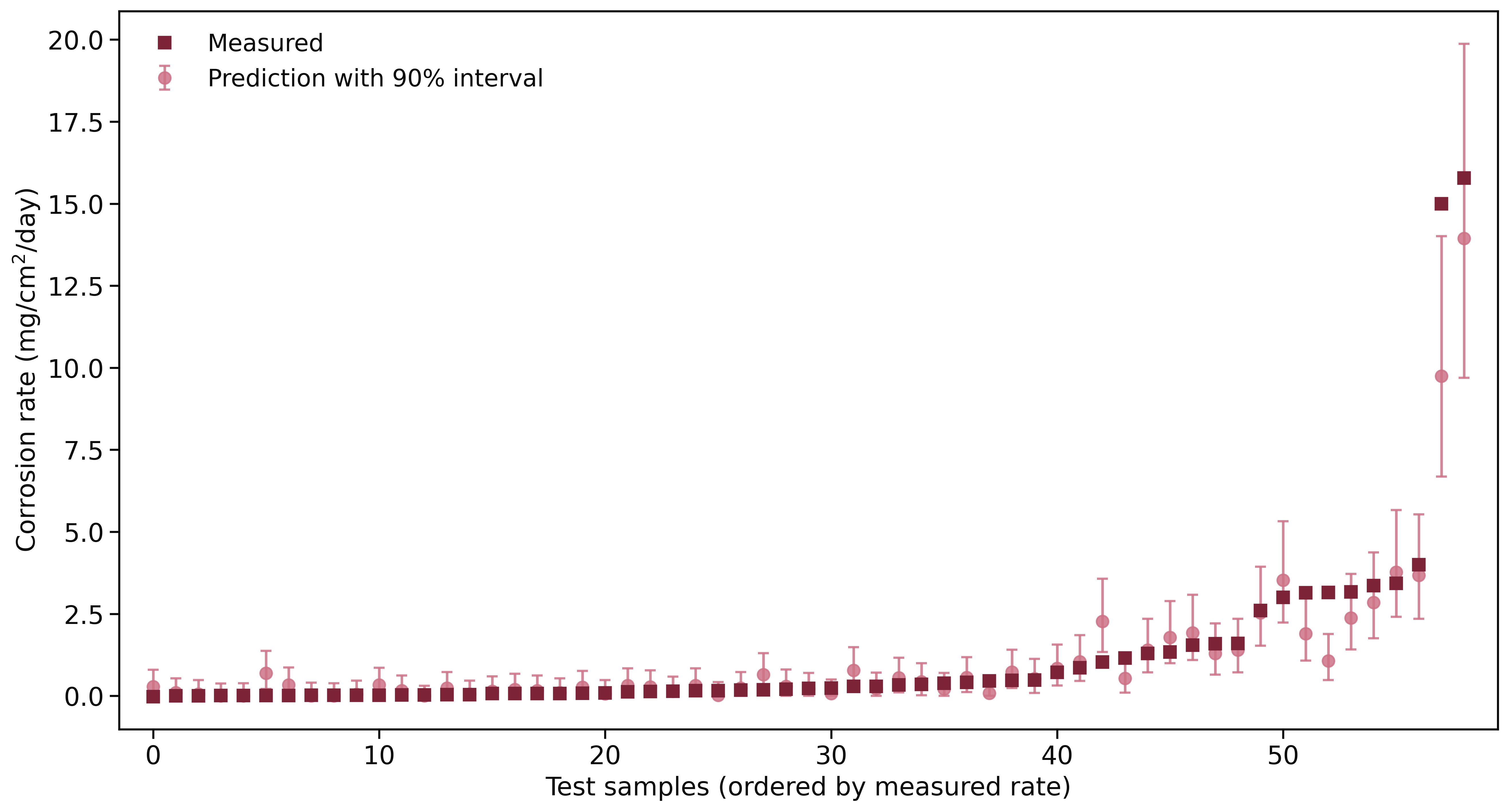}} 
    \caption{Measured corrosion rates and ensemble predictions with 90\% conformal prediction intervals. }
    \label{fig:fg7}
   \end{figure}

However, point predictions provide no direct measure of the uncertainty associated with an individual corrosion-rate estimate. This is particularly important for high-rate observations, where the parity plots show greater prediction scatter. Therefore, the conformal prediction intervals were calculated for the ensemble to evaluate whether the uncertainty bounds could reflect this variation. \Cref{fig:fg7} shows the measured values and the corresponding predictions with 90\% intervals for a representative held-out partition, with samples ordered by the measured corrosion rate. The intervals remain relatively narrow through the densely populated low-rate region and broaden progressively for higher corrosion rates. The measured value falls within its interval for 52 of the 59 records, including many of the observations that appear as outliers in the parity plot.

\begin{figure}[H]
    \centering
    \subfigure[]{\includegraphics[width=0.46\linewidth]{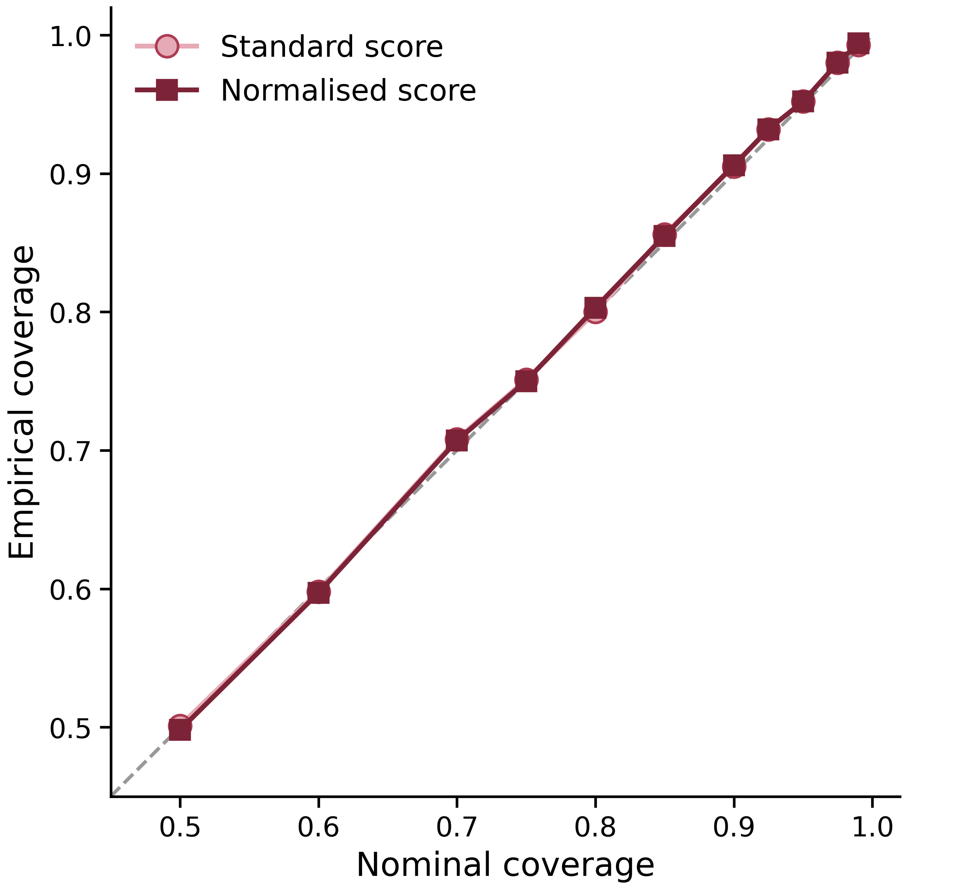}}
    \hspace{1cm}
	\subfigure[]{\includegraphics[width=0.44\linewidth]{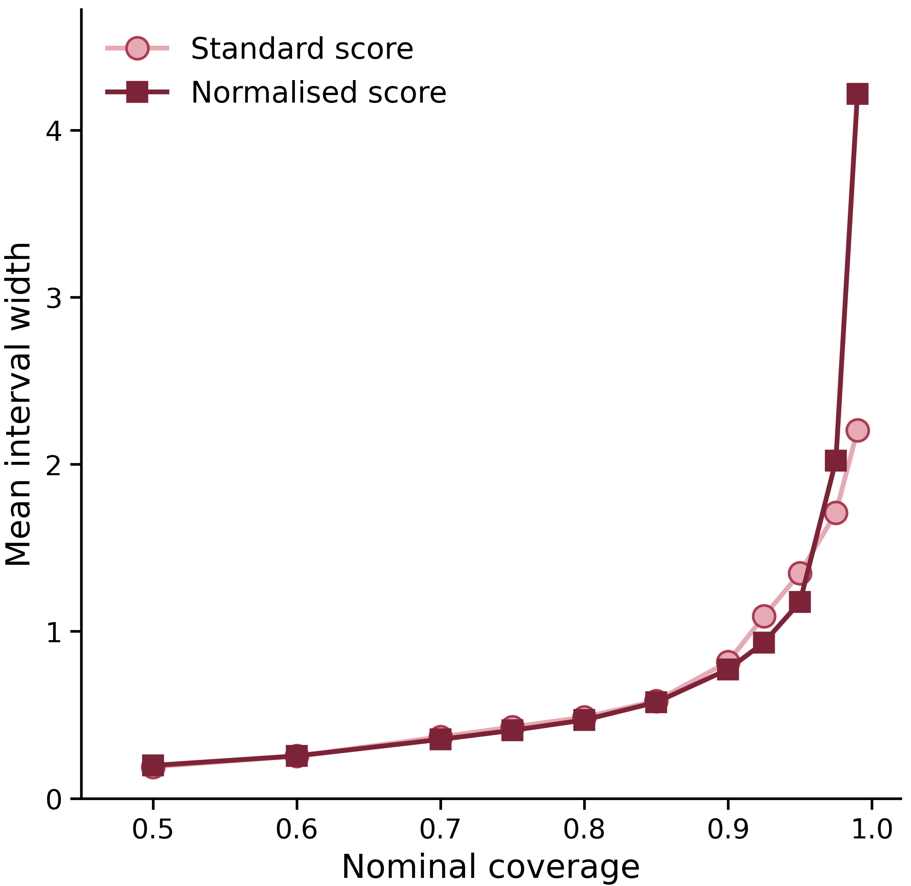}} 
    \caption{Conformal prediction interval calibration showing (a) empirical coverage and (b) interval width across nominal coverage levels. }
    \label{fig:fg8}
   \end{figure}

Because the coverage obtained from a single test partition can depend on the observations included, the calibration behavior was also evaluated using the complete set of out-of-fold predictions following the procedure described in \cref{subsection:2.5}. For the calibration analysis, predictions were taken from the first repetition of the five-fold cross-validation, in which each observation is held out exactly once and, therefore, predicted by a model that had not been trained on that observation. To assess the consistency of this result, repeating the analysis using the second repetition or the average of the two out-of-fold predictions changed the empirical coverage at the 0.80, 0.90, and 0.95 nominal levels by less than 0.003. As shown in \cref{fig:fg8} (a), the empirical coverage closely follows the nominal coverage. The uncertainty bounds therefore provide a calibrated measure of prediction reliability rather than simply reflecting the scatter of the point predictions.

The intervals are also narrow enough to be informative; at the 90\% level the mean half-width is 0.41 on the target. As expected, a higher coverage is accompanied by an increase in interval width (\cref{fig:fg8} (b)). With model performance and uncertainty now evaluated, the following sections investigate how individual descriptors contribute to the model predictions.

\subsection{Influence of salt preparation representation on model performance}

Salt purity strongly influences corrosion in molten chlorides, yet it is often described qualitatively in the literature rather than reported as a directly measured quantity. Previous data-aggregation studies have represented salt conditions using a binary classification, such as purified/unpurified \cite{raiman2018aggregation}, while individual experimental studies have also distinguished low, moderate and high levels of salt purity \cite{kurley2019enabling}. Although convenient, this approach treats substantially different preparation routes as equivalent. For example, thermal drying, inert-gas treatment, Mg purification, and chemical sparging may all be placed in the same purified category despite producing very different salt conditions. Raiman and Lee \cite{raiman2018aggregation} similarly noted that such binary classifications were necessarily somewhat arbitrary because the purification and handling procedures varied widely among studies and quantitative impurity measurements were often unavailable. We therefore constructed the Chloride Salt Preparation Index (CSPI), which scores the procedure reported by each study against a fixed rubric so that the same reported treatment receives the same value irrespective of the study and the corrosion rate that the study measured. The index averages three equally weighted components, a preparation score, an atmosphere-control score, and a pre-exposure impurity score assigned from reported water and oxygen concentrations. Records in which an oxidizing species was deliberately added are excluded, and the full rubric is given in \cref{app:cspi}. 

Because component scores, equal weighting, and the impurity score are assigned using rubric values, the sensitivity of model performance to these choices was also examined at the salt-batch level, so that all records sharing a prepared salt were adjusted together. Each alternative rubric was applied as the change it produced in the component scores of every batch, as scored from the source reports. Changing any component weight by $\pm 0.10$ altered a batch score by at most 0.085 and gave MAE between 0.411 and 0.421 mg\,cm$^{-2}$\,day$^{-1}$, against 0.410 mg\,cm$^{-2}$\,day$^{-1}$ for the reference rubric, and a batch-level perturbation of $\pm 0.10$ raised MAE only to 0.419 mg\,cm$^{-2}$\,day$^{-1}$. Replacing the impurity score for the moderate class ($P_{imp} = 0.30$) by 0.05, 0.60, or 0.90 for batches without an impurity analysis, or omitting the impurity component for these batches, shifted the batch scores by up to 0.21 and gave MAE between 0.413 and 0.417 mg\,cm$^{-2}$\,day$^{-1}$. In every case, the graded index approach retained a lower MAE than both the binary purified/unpurified label and its omission.

\begin{figure}[H]
    \centering
	\subfigure[]{\includegraphics[width=0.45\linewidth, height= 8.5 cm, keepaspectratio]{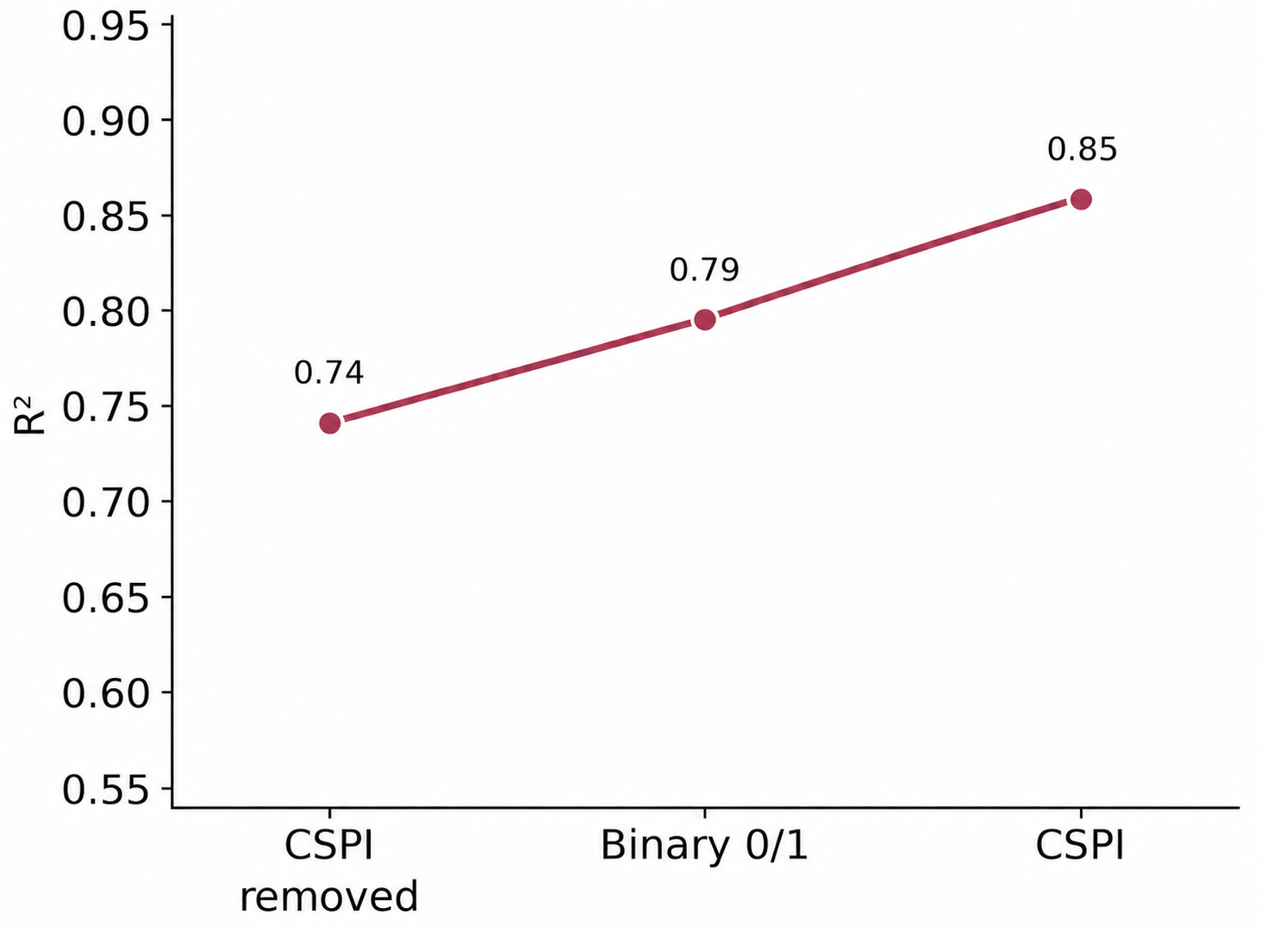}} \hspace{1cm}
    \subfigure[]{\includegraphics[width=0.45\linewidth, height= 8.5 cm, keepaspectratio]{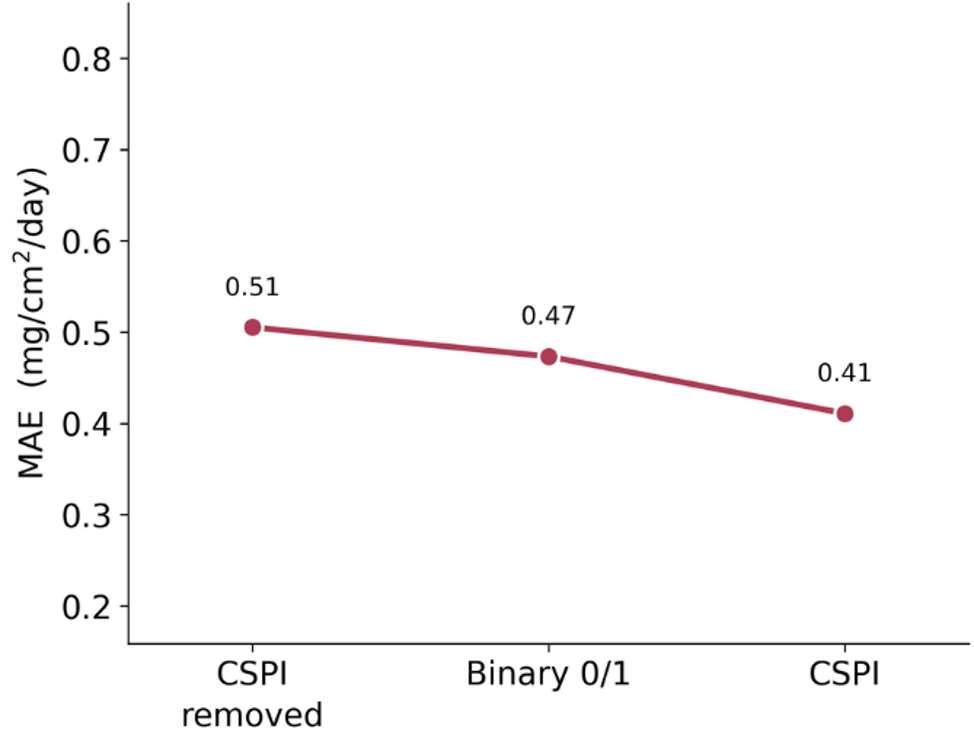}}
    \caption{Comparison of model performance using different salt-purification representations: (a) \(R^2\) and (b) MAE for no purification feature, binary 0/1 classification, and the CSPI descriptor.}
    \label{fig:fg12}
   \end{figure}

Three model configurations were evaluated to isolate the contribution of the CSPI representation. In the first configuration, we trained the model without any purification feature. Later, each record was classified as purified or impure following the criteria of Raiman and Lee \cite{raiman2018aggregation}, with impure salt coded 0 and purified salt coded 1. In the third configuration, the continuous CSPI value was used instead of the binary 0/1 variable. All other parts of the modeling workflow were kept unchanged, including the remaining input features, preprocessing, scaling, target transformation, ensemble structure, and cross-validation folds. Each configuration was therefore evaluated on the same held-out records, allowing a direct fold-by-fold comparison of the effect of the purification representation. The results are shown in \cref{fig:fg12}.

As seen in \cref{fig:fg12}, the absence of the salt-purification descriptor had a clear negative effect on model performance. When purification was represented only as a binary variable (0 = impure, 1 = purified), performance improved. Continuous CSPI gave the best results, reaching \(R^2=0.85\) and an MAE of 0.41 mg cm\(^{-2}\) day\(^{-1}\). This trend shows that a simple purified/impure label captures some useful information, but a graded description of the salt preparation retains additional variation that is relevant to corrosion prediction. The improvement was observed across the held-out folds. CSPI was also the highest-ranked descriptor in the fitted model, exceeding the importance of any individual alloying element or salt constituent.

As an additional control, randomly reassigning CSPI values among the records resulted in poorer performance than removing the purification feature entirely. This indicates that the observed improvement is associated with the information represented by CSPI rather than simply adding another numerical variable to the model. Nevertheless, CSPI reflects the salt condition as reported in each study, rather than a direct measurement of absolute purity. A low CSPI may therefore represent either limited salt preparation or limited reporting of the procedure, and the index should be interpreted as a graded descriptor of reported salt-preparation quality rather than as a direct purity measurement.

\subsection{Model Interpretation and Global Sensitivity Analysis}

To investigate how much of the variation in the predicted corrosion rate could be attributed to each group of inputs, a variance-based global sensitivity analysis was carried out using Sobol' indices. The first-order index of an input measures the share of the output variance that the input explains on its own, while the total index adds any contribution it makes through interaction with other inputs.

A conventional Sobol' analysis samples every input independently across its range. That approach is not suitable here because the twelve coupon elements and the seven salt constituents are each constrained to sum to 100\%, and independent sampling would produce compositions that cannot exist. The inputs were therefore arranged into seven groups: coupon chemistry, salt chemistry, container, temperature, exposure time, CSPI and V/SA. Each group was sampled as a complete vector drawn from the records in the dataset, so that every sampled alloy and every sampled melt corresponds to one that was actually tested. The first-order indices were estimated with the Saltelli estimator and the total indices with the Jansen estimator \cite{saltelli2010, jansen1999}, from $N = 4096$ base samples ($N(k+2) = 36{,}864$ model evaluations for $k = 7$ groups). Confidence intervals were obtained from 400 bootstrap resamples of the base samples.

\begin{figure}[H]
    \centering
    \includegraphics[width=0.95\linewidth, height=7 cm, keepaspectratio]{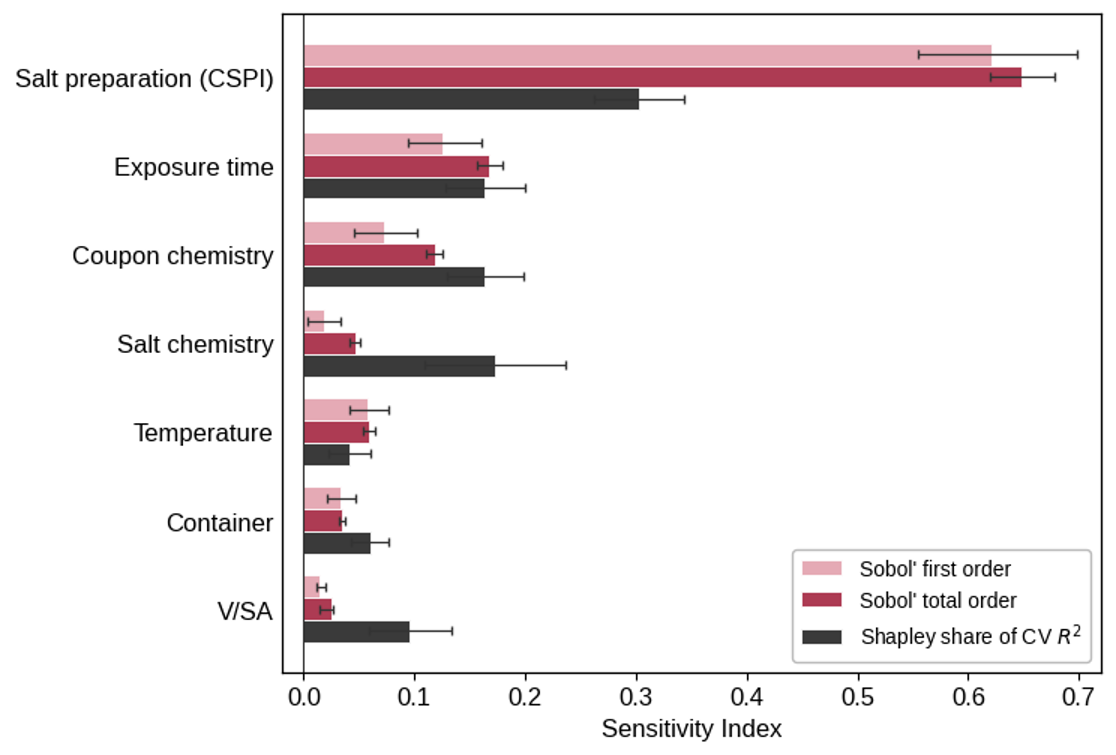}
    \caption{Comparison of grouped Sobol' sensitivity indices and Shapley contributions to cross-validated \(R^2\) for the seven input groups. Error bars represent 95\% bootstrap confidence intervals..}
    \label{fig:fg10}
\end{figure}

Because each group is sampled independently of the others, the indices describe the sensitivity of the fitted model under that sampling distribution. In the compiled literature, the groups are not independent. CSPI, salt chemistry, alloy family, and test conditions vary from study to study. As a dependency-aware check, the ensemble was refitted on every subset of the seven groups, and the cross-validated $R^2$ on the target was apportioned among the groups by their Shapley values \cite{lipovetsky2001}. This analysis operates on the observed records rather than independently generated combinations of the input groups, the correlations present in the literature dataset are retained. The two analyses are compared in \cref{fig:fg10}.

Under independent group sampling, CSPI is the dominant input, accounting for 62\% of the variance of the model output alone and 65\% when its interactions are included. Exposure time is the second most influential group, followed by coupon chemistry, which taken as a group of twelve elements explains less of the variance than exposure time alone. The corrosion rate slows as the exposure continues, longer tests report lower average rates, and the model draws on the exposure time accordingly. Temperature, container, salt chemistry and V/SA each account for 6\% or less.

When the dependence among the groups is retained, CSPI remains the largest contributor and its share of predicted performance was 30\%. Salt chemistry, exposure time, coupon chemistry and V/SA then carry comparable shares, and removing salt chemistry from the model reduces the cross-validated $R^2$ by 0.12. The two analyses agree that salt preparation and exposure time are among the leading influences on the predicted rate. The difference between them shows that part of the influence attributed to CSPI under independent sampling reflects the information it shares with salt chemistry and with study-level practice in the compiled data.

\begin{figure}[H]
    \centering
      \hspace{3cm} {\includegraphics[width=0.75\linewidth, height= 8.5 cm, keepaspectratio]{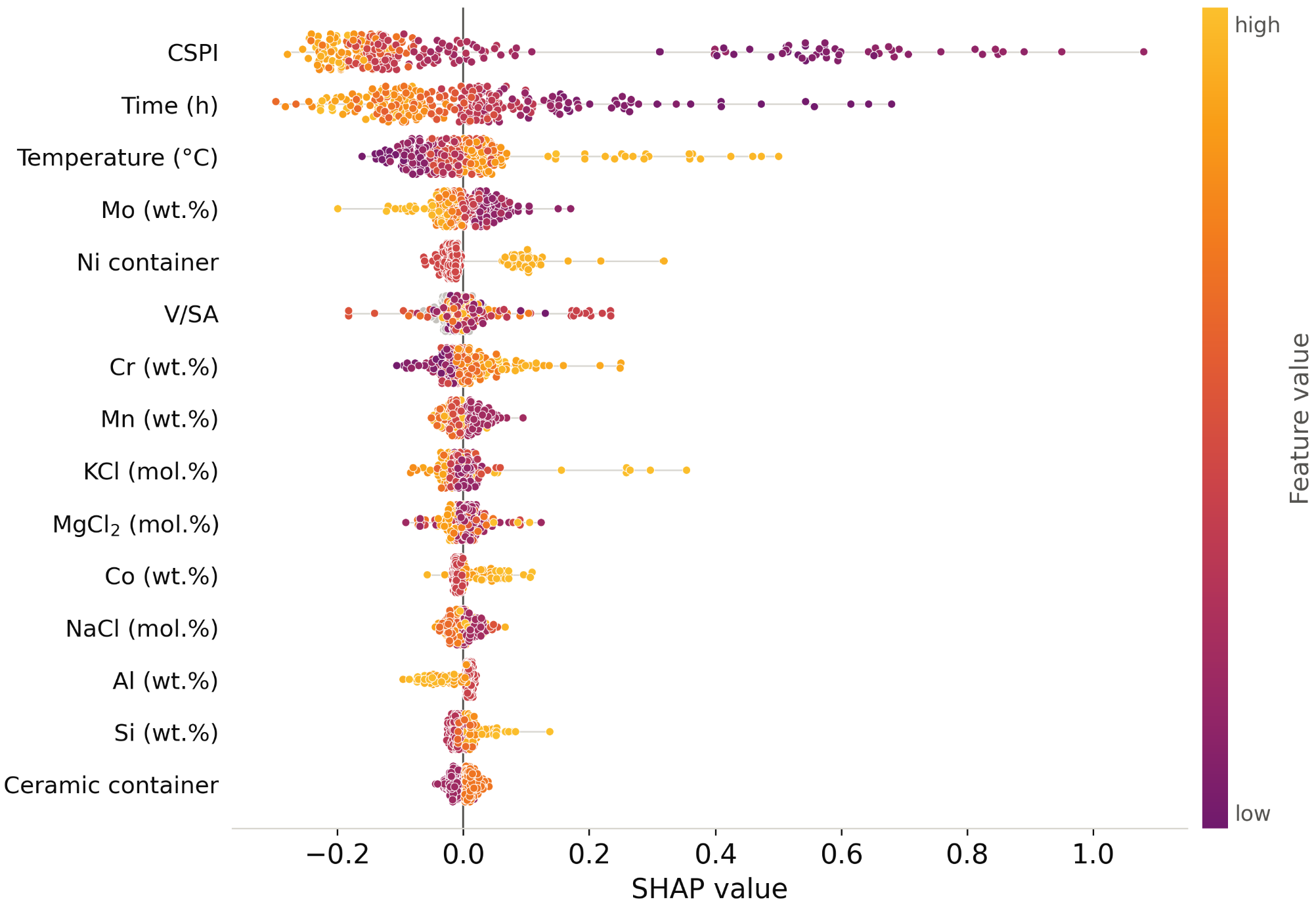}}
    \caption{SHAP-based ranking of the model inputs and their contribution to the predicted corrosion rate.}
    \label{fig:fg11}
   \end{figure}

While the Sobol analysis identifies the contribution of each input group to the overall variance in the model prediction, it does not reveal the influence of the individual variables within those groups on the corrosion rate. SHAP analysis was therefore used to provide a feature-level interpretation of the fitted model. The features in \cref{fig:fg11} are ordered by their mean absolute SHAP value and therefore by their overall influence on the model.

The SHAP results reinforce the trends identified by the grouped Sobol analysis. CSPI has the largest overall influence on the model, followed by exposure time and temperature. Higher CSPI values are concentrated mainly at negative SHAP
values, showing that better-controlled salt conditions generally reduce the predicted corrosion rate, whereas low CSPI values tend to increase it. Exposure time shows a similar trend, with longer exposures associated mainly with lower predicted rates. As exposure continues, the system gradually approaches a quasi-steady state where the corrosion rate decreases or saturates, often attributed to the depletion of reactive impurities and chemical equilibration between the metal and salt \cite{sridharan2013corrosion,pillai2021first}. Temperature shows the opposite model response, with higher temperatures generally associated with higher predicted corrosion rates. This trend is consistent with the reported effects of temperature on chromium transport, chloride solubility, and corrosion kinetics\cite{pillai2021first,ma2022comparative}.

Among the alloying elements, Mo and Cr show the clearest directional effects. Higher Mo contents generally shift the prediction towards lower corrosion rates, whereas higher Cr contents generally shift the prediction toward higher corrosion rates, consistent with the preferential dissolution of Cr in molten chlorides \cite{sridharan2013corrosion,rippy2023predicting}. The presence of a Ni container also produces predominantly positive SHAP values, indicating a
measurable influence of the test environment. The V/SA, individual salt
constituents, and remaining alloying elements have smaller and more
distributed contributions, suggesting that their effects depend more strongly on the surrounding composition and experimental conditions.

\section{Conclusion}
This work demonstrates that a data-driven framework can capture nonlinear associations among alloy composition, salt chemistry, and test conditions in static molten chloride corrosion data while providing quantified prediction uncertainty. It also establishes a methodological template for compatibility assessment in extreme environments. Based on the dataset compilation, model development, and interpretation analysis, the following conclusions can be drawn.

\begin{itemize}
    \item Tree-based ensembles predict the corrosion rate substantially better than kernel and neural-network models on this dataset. A blend of four such learners was more consistent between folds than any individual learner.
    
    \item The model identified salt preparation as the most influential predictive descriptor. Representing salt preparation using the graded CSPI rather than a binary purified/unpurified classification improved predictive performance, indicating that differences in reported salt treatment, atmosphere control, and impurity condition contain useful information for corrosion-rate prediction.

    \item The global sensitivity analysis identified salt preparation, exposure time, and coupon chemistry as the main contributors to variation in the model predictions, while the SHAP analysis ranked the contributions of individual descriptors within these groups. The model captures trends that are consistent with the known molten-salt corrosion behavior. The higher Mo content is associated with lower predicted corrosion rates, whereas higher Cr content is associated with higher rates, consistent with the preferential dissolution of Cr in chloride salts. Longer exposure times are associated with lower average corrosion rates, while higher temperatures tend to increase the predicted rate.
    
    \item The addition of the salt volume-to-surface-area ratio ($V/SA$) improved the  predictive performance, increasing $R^2$ from 0.81 to 0.85 and reducing MAE from 0.43 to 0.41~mg\,cm$^{-2}$\,day$^{-1}$. 

   \item  Uncertainty quantification showed that the conformal prediction intervals were well calibrated across the evaluated confidence levels. The empirical coverage closely followed the nominal coverage. This analysis provides additional information beyond the point predictions by showing the level of uncertainty associated with individual corrosion-rate estimates.
  \end{itemize}
 
A significant challenge throughout this work was the inconsistency of reporting in the historical literature. Although the models made effective use of the
available descriptors, several parameters known to influence corrosion kinetics were frequently absent or inconsistently documented. Consistent reporting of the salt purification procedure, the salt volume-to-wetted-surface-area ratio,
the in-situ redox potential and the coupon microstructure would reduce the uncertainty of future predictive models more effectively than additional records collected under present reporting practice.

More fundamentally, static tests cannot reproduce all phenomena present in operating systems, particularly thermal-gradient mass transfer, forced convection and
flow-dependent transport. The model should therefore be regarded as a foundation for data-driven analysis of molten-salt corrosion rather than as a substitute for dynamic loop testing. Establishing this static-data framework is
a necessary first step towards models that incorporate flow velocity, thermal gradients, mass transport and mechanical loading, and that can relate static corrosion data to service-relevant molten-salt environments systematically.

\section*{Acknowledgment}
The author gratefully acknowledge Dr. Tasnim Hassan for his support, insightful suggestions, and constructive technical discussions that helped strengthen the development and interpretation of this work.


\appendix
\crefalias{section}{appendix}
\section{Chloride Salt Preparation Index (CSPI)}
\label{app:cspi}

The Chloride Salt Preparation Index (CSPI), bounded on $[0,1]$, encodes the
\textit{reported initial condition} of a molten chloride salt as a single
dataset feature. It combines the reported purification method, atmosphere
control, and pre-exposure impurity condition. CSPI is not a measurement or
percentage of absolute salt purity. Corrosion rate, alloy, temperature,
exposure duration, salt chemistry, containment, flow condition, and
intentionally added redox species are treated as independent features and
shall not influence the index.

\subsection{Nomenclature}

\begin{table}[H]
\centering
\small
\begin{tabular}{lll}
\hline
\textbf{Symbol} & \textbf{Definition} & \textbf{Domain} \\
\hline
$P_{\mathrm{method}}$ & Salt-preparation score & $[0.10,0.90]$ \\
$P_{\mathrm{atm}}$    & Atmosphere-control score & $[0.10,0.95]$ \\
$P_{\mathrm{imp}}$    & Pre-exposure impurity score & $[0.05,0.90]$ \\
CSPI                  & Chloride Salt Preparation Index & $[0,1]$ \\
\hline
\end{tabular}
\end{table}

\subsection{Procedure}

\subsubsection{Step 1: Applicability}

The observation is retained with $\mathrm{CSPI}=\mathrm{NA}$ if:

\begin{enumerate}
    \item an oxidizing or redox-controlling species is intentionally added
    as the experimental condition;
    \item an oxidizing gas is intentionally supplied during exposure; or
    \item the melt is not predominantly chloride based.
\end{enumerate}

\subsubsection{Step 2: Assign $P_{\mathrm{method}}$}

\begin{table}[htbp]
\centering
\small
\begin{tabularx}{\linewidth}{
    >{\centering\arraybackslash}p{1.0cm}
    X
    >{\centering\arraybackslash}p{1.5cm}
}
\toprule
\textbf{Class} & \textbf{Reported preparation} & 
\textbf{$P_{\mathrm{method}}$} \\
\midrule

0 &
As received, mixed/melted only, or ``purified'' without a stated method
& 0.10 \\

1 &
Thermal or physical treatment: drying, vacuum drying,
high-temperature dehydration, or filtration
& 0.50 \\

2 &
Chemical treatment: reducing metal, carbochlorination, CCl$_4$,
HCl/Cl$_2$, Ar--H$_2$, distillation, or electrochemical purification
& 0.90 \\

\bottomrule
\end{tabularx}
\end{table}

\noindent Three broad classes are used primarily because treatment temperature,
duration, and other procedural details are not reported consistently enough
across the literature to justify finer numerical resolution.
%
%



\subsubsection{Step 3: Assign $P_{\mathrm{atm}}$}

\begin{table}[H]
\centering
\small
\begin{tabular}{clc}
\hline
\textbf{Class} & \textbf{Reported condition} & $P_{\mathrm{atm}}$ \\
\hline
0 & Air, or atmosphere not reported & 0.10 \\
1 & Inert gas; H$_2$O/O$_2$ not quantified & 0.60 \\
2 & Controlled atmosphere with quantified H$_2$O/O$_2$ & 0.80 \\
3 & Sealed system with quantified low H$_2$O/O$_2$ ($\leq$ 1 ppm) atmosphere & 0.95 \\
\hline
\end{tabular}
\end{table}

\subsubsection{Step 4: Assign $P_{\mathrm{imp}}$}
H\(_2\)O and oxygen-related impurities were selected as the primary impurity indicators because they are among the most commonly reported contaminants in molten chloride studies and are directly associated with hydrolysis, oxychloride formation, and changes in salt redox chemistry. The index therefore does not imply that these are the only impurities affecting corrosion. Where pre-exposure H$_2$O and oxygen concentrations are reported, the
impurity condition is classified according to Table~\ref{tab:cspi_impurity}.

\begin{table}[htbp]
\centering
\caption{Classification of pre-exposure salt impurities.}
\label{tab:cspi_impurity}
\small
\begin{tabular}{lccc}
\hline
\textbf{Class} &
\textbf{H$_2$O (ppm)} &
\textbf{O (ppm)} &
$P_{\mathrm{imp}}$ \\
\hline
Very clean & $\leq 5$   & $\leq 30$  & 0.90 \\
Clean      & $\leq 20$  & $\leq 100$ & 0.60 \\
Moderate   & $\leq 100$ & $\leq 500$ & 0.30 \\
High       & $>100$     & $>500$     & 0.05 \\
\hline
\end{tabular}
\end{table}

When both H$_2$O and O are reported, each is classified independently and
the more conservative class is retained:

\begin{equation}
\mathrm{impurity\ class}
=
\min
\left(
\mathrm{class}_{\mathrm{H_2O}},
\mathrm{class}_{\mathrm{O}}
\right).
\label{eq:cspi_impclass}
\end{equation}

\noindent If no pre-exposure impurity analysis is reported, $P_{\mathrm{imp}}$, is assumed to be in the moderate class. This is a prespecified imputation for an
\textit{unverified} salt condition and does not always imply that the salt was measured to contain moderate impurity.

\subsubsection{Step 5: Calculate CSPI}

The three components are combined using equal weighting:

\begin{equation}
\mathrm{CSPI}
=
\frac{
P_{\mathrm{method}}
+
P_{\mathrm{atm}}
+
P_{\mathrm{imp}}
}{3}
\label{eq:cspi}
\end{equation}

Equal weighting is used as a minimum-assumption formulation because the available dataset does not provide sufficient independent evidence to justify optimized component weights.

\clearpage
\bibliographystyle{ieeetr}
\bibliography{Ref}
\end{document}